\documentclass[a4paper, 10pt, conference]{ieeeconf}      % Use this line for a4 paper

\IEEEoverridecommandlockouts                              % This command is only needed if 
\usepackage{graphics} % for pdf, bitmapped graphics files
\usepackage{epsfig} % for postscript graphics files
\usepackage{times} % assumes new font selection scheme installed
\usepackage{amsfonts}
\usepackage{amssymb}  % assumes amsmath package installed

\usepackage{graphicx}
\usepackage{amsmath}
\usepackage{algorithm}
\usepackage{algorithmic}
\usepackage{multirow}
\usepackage{soul}
\usepackage{color}
\usepackage{pifont}
\usepackage{pgfplots}
\usepackage[outline]{contour}% http://ctan.org/pkg/contour
\usepackage{subcaption}
\usepackage{booktabs}
\usepackage{eqparbox}

\usepackage{xspace}
\makeatletter
\DeclareRobustCommand\onedot{\futurelet\@let@token\@onedot}
\def\@onedot{\ifx\@let@token.\else.\null\fi\xspace}

\def\etal{\emph{et al}\onedot}
\makeatother

\title{\LARGE \bf Multiple data sources and domain generalization learning method for road surface defect classification}

\author{Linh Trinh, Ali Anwar, Siegfried Mercelis\\
{Faculty of Applied Engineering, IDLab, University of Antwerp-imec, Belgium} \\
{\tt\small\{linh.trinh, ali.anwar, siegfried.mercelis\}@uantwerpen.be}\\
}

\begin{document}
\pdfoutput=1

\maketitle
\thispagestyle{empty}
\pagestyle{empty}

%%%%%%%%%%%%%%%%%%%%%%%%%%%%%%%%%%%%%%%%%%%%%%%%%%%%%%%%%%%%%%%%%%%%%%%%%%%%%%%%
\begin{abstract}
Roads are an essential mode of transportation, and maintaining them is critical to economic growth and citizen well-being. With the continued advancement of AI, road surface inspection based on camera images has recently been extensively researched and can be performed automatically. However, because almost all of the deep learning methods for detecting road surface defects were optimized for a specific dataset, they are difficult to apply to a new, previously unseen dataset. Furthermore, there is a lack of research on training an efficient model using multiple data sources. In this paper, we propose a method for classifying road surface defects using camera images. In our method, we propose a scheme for dealing with the invariance of multiple data sources while training a model on multiple data sources. Furthermore, we present a domain generalization training algorithm for developing a generalized model that can work with new, completely unseen data sources without requiring model updates. We validate our method using an experiment with six data sources corresponding to six countries from the RDD2022 dataset. The results show that our method can efficiently classify road surface defects on previously unseen data.
\end{abstract}

%%%%%%%%%%%%%%%%%%%%%%%%%%%%%%%%%%%%%%%%%%%%%%%%%%%%%%%%%%%%%%%%%%%%%%%%%%%%%%%%
\section{INTRODUCTION}
Roads play a crucial role in transportation. Research domains especially road engineering study the problem of road surface condition inspection for road maintenance. Periodic road maintenance is an important task, cities and municipalities allocate large sums of budget to repair the defective parts and therefore require automated road defects measurements for cost-effective road maintenance. Multiple criteria exist for evaluating visual surface quality, including the Pavement Condition Index (PCI) \cite{astm2012standard}, Pavement Surface Condition Index (PSCI) \cite{psci}, and Pavement Surface Evaluation Rating (PASER) \cite{paser}. These standards provide measurement instructions for pavement characteristics used in pavement assessments. Nowadays, with the growth and computational support of AI and deep learning solutions, road maintenance can be performed sufficiently accurately and automatically using data collected on the road (with the supporting of compliance of regulations \cite{pp4av,fisheyepp4av}). More specifically, a set of sensors such as a camera, lidar, radar, or GPS can be mounted on a vehicle and driven along the road to collect data, after which a developed machine learning model is used to determine the road surface conditions for assessment. With the focus on these tasks, there has been a strong emphasis on a large data-driven approach to building them. 
In addition, various recent studies on datasets and methods from a variety of countries have recently been announced, providing visual data for road surface condition assessments using AI/deep learning methods, such as LTPP-FHWA in the USA \cite{data_ltpp}, GAPs in Germany \cite{data_GAPs}, CFD from China \cite{cfd_china}, and RDD in China, Japan, India, the Czech Republic, Norway, and United States \cite{RDD19,RDD20,RDD22}. This provides a strong evidence that it is an appealing topic for research on transportation.

There has been a growing interest in studying road surface inspection within computer vision community in recent years. Several recent approaches have been proposed in the field of road surface classification. The majority of the proposed methods rely on the architecture of Convolutional Neural Networks (CNN), as exemplified by \cite{road_surface_CNN,road_surface_RCNet,road_surface_condition,road_surface_cnn2,road_surface_dcnn}. These methods involve proposing the architecture through the design or modification of neural network structures. Alternative approaches utilize a pretrained backbone on ImageNet dataset, such as ResNet50, InceptionNetV3 \cite{nolte2018assessment}, or ConvNeXt \cite{road_surface_fusion} to construct the classification. Another approach involves integrating Convolutional Neural Networks (CNN) with other neural network architectures, such as Transformer \cite{road_surface_cnn-transformer} and Long Short-Term Memory (LSTM) \cite{road_surface_iv}. Alternatively, certain approaches incorporate data fusion or feature fusion techniques to improve the classification task, as demonstrated in studies such as \cite{road_surface_fusion, road_surface_shadow, road_surface_cnn-transformer}. 
In the classification of road conditions, several approaches rely on Convolutional Neural Networks (CNN) to construct classifiers, as demonstrated in the works of \cite{rtk_road_quality,method_bleeding,method_cnn_soict,method_crack_pave_vn}. The approach of constructing a classifier using a pretrained backbone network has been extensively researched, with notable examples including VGG16, MobileNetV2 \cite{method_cnn_2017,method_transfer_learning,method_light_weight_nn}, ResNet18 \cite{method_raveling}, and ResNet101 \cite{method_orthoframes}. Several other efforts targeted the improvement of the inspection of road surface conditions by incorporating additional signals such as wheel-road interactions \cite{method_embedded}, accelerometers \cite{method_sensor}, orthoframes \cite{method_orthoframes}.

However, existing road surface defect classification methods are designed to train models on just one source of data. When data source invariance occurs, these methods may not be applicable to leveraging multiple different data sources for training purposes. Furthermore, because those methods were not trained on this dataset, they may be inefficient when applied to new, previously unseen data sets. In this paper, we take on the context in which multiple data sources are available and how to train a model efficiently on these data sources. Furthermore, we consider the issue of the model's low accuracy on the new, unseen dataset. So we present a method for overcoming both of these challenges.

In summary, our main contributions are as follows: 
\begin{itemize} 
    \item We present our method for classifying road surface defects based on camera images. Our method aims to address two issues: resolving invariance when leveraging multiple data sources for training models, and generalizing the model to work with unseen datasets without training the model on them. Our method consists of two parts: contrastive learning-based training and a new algorithm for domain generalization training.
    \item We conduct extensive experiments to evaluate our method using six data sources from the RDD2022 public road surface defect dataset. The experiment compares the results of our method to recent studies on classification for road surface defect classification, demonstrating that our method is efficient and well-suited to the unseen dataset.
\end{itemize}
We organize the rest of the paper as follows. We present our literature studies on the related works in Section \ref{sec:related_work}. Then, in Section \ref{sec:method}, we present our methodology. Section \ref{sec:exp} discusses our experiment and results. Finally, we conclude our work in Section \ref{sec:conclude}.
%%%%%%%%%%%%%%%%%%%%%%%%%%%%%%%%%%%%%%%%%%%%%%%%%%%%%%%%%%%%%%%%%%%%%%%%%%%%%%%%
\section{RELATED WORKS} \label{sec:related_work}
Several previous studies have focused on deep learning-based methodologies for classifying road surface defects. Li \etal \cite{defect_dcnn@Li_2020} proposed multiple-stage networks composed of CNNs for a classifier model. This method has been verified on a private dataset containing four defect classes: alligator, block, edge, logitudinal, and transverse and reflection cracks.
Nhat \etal \cite{defect_tree_cnn_nn@Nhat_2023} proposed a framework for determining pavement fatigue severity using decision trees, deep neural networks, and CNN. The model was tested on a private dataset containing alligator, block, and minor fatigue area cracks. 
More recently, Mai \etal (2024) proposed a multilabel convolutional neural network to classify asphalt distress. The model was validated using a private dataset that included four distress classes: alligator, block, longitudinal/transverse cracks, and pothole.
According to \cite{defect_dcnn_pavement@Elham_2023}'s literature review, the most commonly used classifiers for transportation are VGG19, VGG16, ResNet50, DenseNet121, and generic CNN.
Chen \etal \cite{defect_mobilenet@Chen_2023} propose a combination of MobileNetV3-large and Convolutional Block Attention Module for crack classification. This study also used focal loss for training. The model was tested on the surface defect of the bridge road.
Silva \etal \cite{defect_vgg16@Silva_2018} propose a transfer learning approach based on VGG16 to classify cracks and non-cracks on concrete roads.
Dais \etal \cite{defect_vgg_resnet_mobilenet@Dais_2021} propose a transfer learning approach using VGG16, MobileNet, and ResNet for non-crack and crack classification.
Ali \etal \cite{defect_transfer_learning@Ali_2021} proposed a customized network based on CNN for crack classification of concrete roads. The model was evaluated against VGG16, VGG19, ResNet50, and InceptionV3.
Liu \etal \cite{defect_infrated_thermo@Liu_2022} proposed a deep learning method and utilized infrared thermography for crack classification in asphalt pavement. Models included MobileNet, DenseNet, EfficientNet, and ResNet.
Dong \etal \cite{defect_few_shot@Dong_2022} propose a deep learning method based on metric learning for multi-target few shot distress classification.
Hsieh \etal \cite{defect_convnet@Hsieh_2021} used ConvNets based on ResNet50 to classify cracks in concrete pavement using 3D images.
Various works also validated their research on privately datasets, such as Gagliardi \etal \cite{method_embedded} propose a method to evaluate pavement quality of road infrastructure using acoustic data of wheel-road interaction. This model is tested on a privately collected dataset with four categories: silence, unknown, good quality, and ruined.
Rateke \etal \cite{rtk_road_quality} propose a CNN network architecture to classify images based on road surface condition. 
Hasanaath \etal \cite{method_transfer_learning} propose a transfer learning method that fine-tunes a classification model using a pre-trained backbone such as VGG16 or MobileNetV2. This model was tested on a private dataset collected in the Kingdom of Saudi Arabia, with four classifications: good, medium, bad, and unpaved.
Chaudhary \etal \cite{method_light_weight_nn} propose a classification model based on pre-trained MobileNetV2 with supplementary layers. The evaluation of an RTK dataset with three main categories: good, normal, and poor quality.
Andrades \etal \cite{method_sensor} propose a classification model based on the self-organizing map (SOM) algorithm. This method, which was tested on private data, consists of several sensors, including an accelerometer and a microcontroller.
Riid \etal \cite{method_orthoframes} propose a classification method based on ConvNets and ResNet101 to detect pavement distress using orthoframes collected by a mobile mapping system in Italy.
Stricker \etal \cite{method_gaps_dataset} use pretrained ResNet34 to improve road condition assessment through transfer learning. This model was tested on privately collected data from Germany, with six classes of road conditions.
Hsieh \etal \cite{method_raveling} developed a transfer learning-based classification model using ResNet18 to detect pavement raveling. The model was tested on a private dataset collected in the United States.
Sajad \etal \cite{method_bleeding} developed a transfer learning-based classification model for pavement bleeding inspection. The method was tested using a private dataset collected in Iran.
Gopalakrishnan \etal \cite{method_cnn_2017} developed a Deep Neural Networks classification using VGG-16 to identify pavement cracks in a dataset collected in the United States.
Nguyen \etal \cite{method_cnn_soict} and Nhat \etal \cite{method_crack_pave_vn} created a CNN architecture to detect pavement cracks on the Vietnamese roads. Li \etal \cite{method_crack_pave_china} created a customized CNN model to detect pavement cracks in China.

To summarize, transfer learning is a common approach of developing classification models that can be tailored to road surface defect classification. However, existing methods were developed for just one source of data, which may be inefficient when training across multiple data sources. Furthermore, these studies are only trained and evaluated on the same dataset, so they cannot be guaranteed to perform well on new data sets. In the following section, we shall explain our methods for tackled both of these challenges.
%%%%%%%%%%%%%%%%%%%%%%%%%%%%%%%%%%%%%%%%%%%%%%%%%%%%%%%%%%%%%%%%%%%%%%%%%%%%%%%%
\section{METHOD} \label{sec:method}
We aim to develop a generalized model using $n$ different available data sources $\mathsf{D}_i,i=1,...,n$. Denote a collection of data sources as $\mathbf{D} = \{\mathsf{D}_1, \mathsf{D}_2,...,\mathsf{D}_n\}$, and $\mathrm{x}_{ij}$ is the $j-$th image of data source $\mathsf{D}_i$. There are $C$ target classes that each sample of $\mathsf{D}_i$ belongs to. We aim to train a generalized classification model that works well on new unseen data sources $\mathbf{D}_{unseen}$ without updating the model on them. We construct a classification model $\mathcal{F}_{\theta}$ that consists of a feature extraction $f_e(;\theta_e)$ ($\theta_e$ is the parameter of $f_e$) and a fully connected layer for generating logits, where $\theta$ represents the model's learnable parameters of whole network $\mathcal{F}_\theta$.
\begin{figure}[ht]
    \centering
    \includegraphics[width=0.51\textwidth]{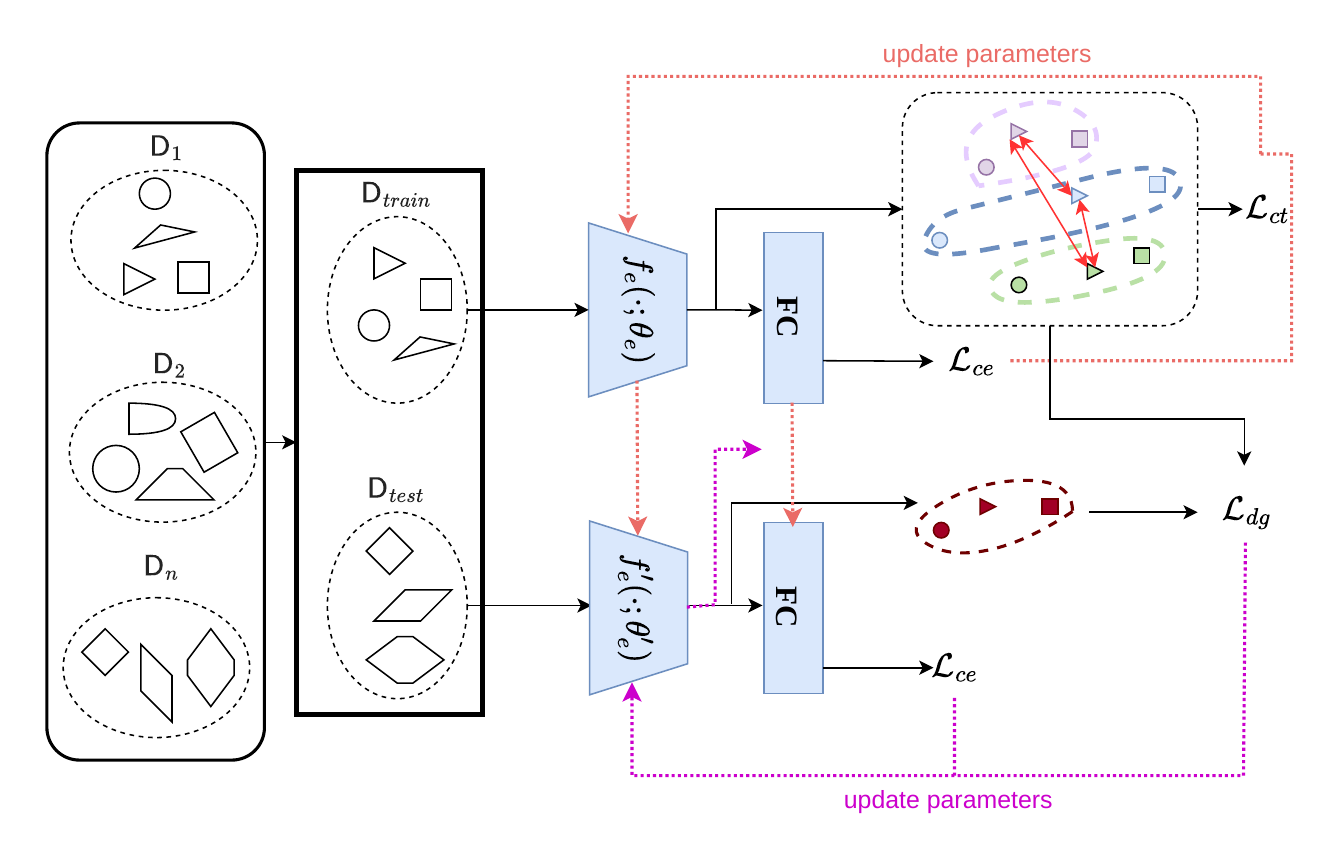}
    \caption{The diagram illustrates our method for training domain generalization and contrastive learning on multiple data sources.}
    \label{fig:fw}
\end{figure}
Figure \ref{fig:fw} illustrates the workflow of our method. We divided the training data source $\mathbf{D}$ into two parts: $\mathsf{D}_{train}$ and $\mathsf{D}_{test}$. In our method, we train model $\mathcal{F}_{\theta}$ on data $\mathsf{D}_{train}$ while optimizing generalization by prioritizing testing data $\mathsf{D}_{test}$. To be more specific, after each step of training a batch of $\mathsf{D}_{train}$, we optimize the generalization process on $\mathsf{D}_{test}$. 

\textbf{Train model on $\mathsf{D}_{train}$.} Training data $\mathsf{D}_{train}$ can be composed from multiple data sources, then we apply contrastive learning \cite{contrastive,contrastive2} to enhance model learning on invariance domain via embedding feature extracted by $f_e(;\theta_e)$. Given a pair of samples ($\mathrm{x}_{ki}, \mathrm{x}_{hj}$), we denote the extracted semantic embedding feature as $\mathrm{p}_{ki}$ and $\mathrm{p}_{hj}$, respectively. In training, we apply contrastive loss to each positive pair ($\mathrm{x}_{ki}, \mathrm{x}_{hj}$) in each training batch of $B_{tr}$, as shown in the equation below.
\begin{equation} \label{eq:consLoss}
\mathcal{L}_{ct}(\mathrm{x}_{ki},\mathrm{x}_{hj})=-\log{\frac{S(\mathrm{x}_{ki},\mathrm{x}_{hj})}{\sum\limits_{\mathrm{x}_{lb}\in B_{tr}} I(\mathrm{x}_{ki},\mathrm{x}_{lb})\cdot S(\mathrm{x}_{ki},\mathrm{x}_{hj})}}
\end{equation}
where $I$ denote an indicator function as below equation:
\begin{equation} \label{eq:identity}
 I(\mathrm{x}_{ki},\mathrm{x}_{lb}) =
    \begin{cases}
      0 & \mathrm{x}_{ki},\mathrm{x}_{lb} \text{ is same class}\\
      1 & \text{otherwise}
    \end{cases}       
\end{equation}
Rather than directly calculating the similarity between two original images $\mathrm{x}_{ki}$ and $\mathrm{x}_{hj}$, we use the corresponding embedded feature. For more information, we use the cosine distance between two features to determine the similarity of two vectors. The following equation determines the similarity of a pair $(\mathrm{x}_{ki},\mathrm{x}_{hj})$:
\begin{equation} \label{eq:similarity}
 S(\mathrm{x}_{ki},\mathrm{x}_{hj})=\exp\left(\frac{SIM(\mathrm{x}_{ki},\mathrm{x}_{hj})}{\tau}\right)
\end{equation}
where $\tau$ is a temperature hyperparameter, and: 
\begin{equation}
    SIM(\mathrm{x}_{ki},\mathrm{x}_{hj})\approx cosine(\mathrm{p}_{ki},\mathrm{p}_{hj})=\frac{\mathrm{p}_{ki}^T\times \mathrm{p}_{hj}}{||\mathrm{p}_{ki}||\times||\mathrm{p}_{hj}||}
\end{equation}
The contrastive loss function is computed for all similar class pairs in batch training for both $(\mathrm{x}_{ki},\mathrm{x}_{hj})$ and $(\mathrm{x}_{hj},\mathrm{x}_{ki})$. Training in this fashion improves the model's ability to explore the invariance of representations, allowing embedding features from the same class to be close together while keeping those from different classes separated.

The training loss on $\mathsf{D}_{train}$ is described below:
\begin{equation} \label{eq:trainLoss}
\mathcal{L}_{tr}=\mathcal{L}_{ce} + \lambda\mathcal{L}_{ct}
\end{equation}
where $\lambda$ is a hyperparameter for controlling the weight of contrastive loss, and $\mathcal{L}_{ce}$ is the cross-entropy loss for multi-label classification task.

\textbf{Domain generalization training.} We propose a strategy for training generalized models to work with unseen datasets. Our method was inspired by recent domain generalization techniques \cite{dg_Guo@2024,dg_Faraki@2021,dg_Guo@2020,dg_Zhao@2021}. The main generalization training of these methods is to align the distribution of features on meta tests with the distribution of features in the meta domain. Unlike most of these techniques, which deal with face recognition and then design metric learning based on triplet loss, we design a new generalization training by optimizing the alignment of feature distribution of testing data source $\mathsf{D}_{test}$ to training data source $\mathsf{D}_{train}$, ulitizing Mahalanobis distance from a point to a distribution. For the distribution of the differential feature vector on the training data source $\mathsf{D}_{train}$, we compute the covariance matrix $\Xi$ and the mean vector $\mu$. We compute the covariance matrix on the entire training data $\mathsf{D}_{train}$ for each class $c\in C$ using the following equation:
\begin{equation}
    \Xi_c=\frac{1}{N}\sum_{i=1}^{N}\left( \mathrm{z}_i - \mu_c \right)\left( \mathrm{z}_i - \mu_c \right)^T
\end{equation}
where $N$ is the number of sample pair of class $c$ $\mathrm{z}_i$ is a different vector of $i-$th pair in this $N$ pair, and $\mathrm{z}_i=f_e(\mathrm{x}_{pg};\theta_e)-f_e(\mathrm{x}_{qh};\theta_e)$ where training pair $\mathrm{x}_{pg}, \mathrm{x}_{qh}$ is the same class $c$, and $\mu_c$ is the mean vector of differential distribution, which is $\mu_c = \textit{mean}(\mathrm{z})$.
After that, with each differential feature vector of testing meta $\mathsf{D}_{test}$, we compute its distance to the distribution of $\mathsf{D}_{train}$ using the Mahalanobis distance as shown below:
\begin{equation}
    \mathcal{M}(\mathrm{z}_i; \Xi_c, \mu_c) = \sqrt{(\mathrm{z}_i-\mu_c)^T\Xi_c(\mathrm{z}_i-\mu_c)}
\end{equation}
Finally, we use $\mathsf{D}_{test}$ to train domain generalization using a loss function as shown in the equation below:
\begin{equation}
    \mathcal{L}_{dg}(\mathsf{B}_{te}; \mathcal{F}_{\theta'})=\frac{1}{C}\sum_{c\in C}\frac{1}{N_c}\sum_{i=1}^{N_c}\mathcal{M}(\mathrm{z}_i; \Xi_c, \mu_c)
\end{equation}
where $\mathrm{z}_i = f_e(\mathrm{x}_{lk}; \theta')-f_e(\mathrm{x}_{gh}; \theta')$ with $\mathrm{x}_{lk}$ and $\mathrm{x}_{lk}$ in $B_{te}$ and same class $c$; $\mathcal{M}(\mathrm{z}_i; \Xi_c, \mu_c)$ is the Mahalanobis distance from a point $\mathrm{z}_i$ to distribution $\mathcal{N}(\mu_c,\Xi_c)$.

The algorithm \ref{al:train} show a detailed description of our method's model training process.
\begin{algorithm}[ht]
\caption{Training domain generalization model}
\label{al:train}
    \hspace*{\algorithmicindent} \textbf{Input:} $n$ data source $\mathbf{D} = \{\mathsf{D}_1, \mathsf{D}_2,...,\mathsf{D}_n\}$  \\
    \hspace*{\algorithmicindent} \textbf{Output:} trained model $\mathcal{F}_{\theta}$
    \begin{algorithmic}[1] %[1] enables line numbers
        \FOR{$i=1,2,...,N_{epoch}$}
            \FOR{Each $\mathsf{D}_{test} \in \mathbf{D}$}
                \STATE $\chi \leftarrow \textit{None}$
                \STATE $\mathsf{D}_{train} = \mathbf{D} \setminus \mathsf{D}_{test}$
                \FOR{Sample batch $B_{te}$ in $\mathsf{D}_{test}$}
                    \STATE Compute $\Xi,\mu=\{\Xi_c\},\{\mu_c\}, \forall c\in C$ on $\mathsf{D}_{train}$
                    \FOR{Sample batch $\mathsf{B}_{tr}$ from $\mathsf{D}_{train}$}
                        \STATE $\mathcal{L}_{tr}\leftarrow\mathcal{L}_{ct}(\mathsf{B}_{tr}; \mathcal{F}_{\theta}) + \mathcal{L}_{ce}(\mathsf{B}_{tr}; \mathcal{F}_{\theta})$
                        \STATE Update weight $\mathcal{F}_{\theta'}\leftarrow\mathcal{L}_{tr}.\textit{backward()}$
                    \ENDFOR
                    \STATE $\mathcal{L}_{te} \leftarrow \mathcal{L}_{ce}(\mathsf{B}_{te};\mathcal{F}_{\theta'}) + \mathcal{L}_{dg}(\Xi, \mathsf{B}_{te}; \mathcal{F}_{\theta'})$
                    \STATE $\chi \leftarrow \chi + \gamma\nabla_{\theta'}\mathcal{L}_{te}$
                \ENDFOR
            \STATE Update weight of $\mathcal{F}$: $\theta'' \leftarrow \theta' -\eta\times\chi$
            \ENDFOR
         \ENDFOR
        \STATE \textbf{return} optimal $\mathcal{F}_{\theta^*}$
    \end{algorithmic}
\end{algorithm}

%%%%%%%%%%%%%%%%%%%%%%%%%%%%%%%%%%%%%%%%%%%%%%%%%%%%%%%%%%%%%%%%%%%%%%%%%%%%%%%%
\section{EXPERIMENT} \label{sec:exp}
\subsection{Settings}
\textbf{Dataset.}
In this experiment, we use RDD2022 \cite{rdd2022_1,rdd2022_4,rdd2022_5}, a public data set for road surface defect study, which data was collected from six countries: China, Japan, Czech, India, Norway, and the United States. We use six data sources for this dataset, which correspond to six countries: China MotorBike, Czech, India, Japan, Norway, and the United States. We crop the bounding boxes from the original dataset based on the annotations provided by the dataset. For consistencies in all six datasets, we retain bounding boxes belonging to four defect classes: longitudinal crack (D00), transverse crack (D10), alligator crack (D20), and a combination of rutting bump, separation cracks, and pothole (D40). We remove small bounding boxes with squared sizes less than 400 $\textit{pixel}^2$. The prepared dataset for our experiment is summarized in the table \ref{tab:table_dataset} and are available here\footnote{https://tinyurl.com/mr45jfhn}.
\begin{table}[ht]
\centering
\caption{Statistic of preprocessed data from RDD2022 for our experiment.}
\label{tab:table_dataset}
\begin{tabular}{|lcccc|c|}
\hline
\multicolumn{1}{|l|}{\multirow{2}{*}{\textbf{Data source}}} & \multicolumn{4}{c|}{\textbf{Defect class}}                                                                               & \multirow{2}{*}{\textbf{SUM}} \\ \cline{2-5}
\multicolumn{1}{|l|}{}                                      & \multicolumn{1}{c|}{\textbf{D00}} & \multicolumn{1}{c|}{\textbf{D10}} & \multicolumn{1}{c|}{\textbf{D20}} & \textbf{D40} &                               \\ \hline
\multicolumn{1}{|l|}{China MotorBike}                       & \multicolumn{1}{c|}{1,419}         & \multicolumn{1}{c|}{738}          & \multicolumn{1}{c|}{486}          & 164          & 2,807                          \\ \hline
\multicolumn{1}{|l|}{Czech}                                 & \multicolumn{1}{c|}{764}          & \multicolumn{1}{c|}{300}          & \multicolumn{1}{c|}{129}          & 154          & 1,347                          \\ \hline
\multicolumn{1}{|l|}{India}                                 & \multicolumn{1}{c|}{1,109}         & \multicolumn{1}{c|}{60}           & \multicolumn{1}{c|}{1,758}         & 1,530         & 4,457                          \\ \hline
\multicolumn{1}{|l|}{Japan}                                 & \multicolumn{1}{c|}{2,797}         & \multicolumn{1}{c|}{2,240}         & \multicolumn{1}{c|}{4,714}         & 1,390         & 11,141                         \\ \hline
\multicolumn{1}{|l|}{Norway}                                & \multicolumn{1}{c|}{2,518}         & \multicolumn{1}{c|}{1,006}         & \multicolumn{1}{c|}{322}          & 256          & 4,102                          \\ \hline
\multicolumn{1}{|l|}{United States}                         & \multicolumn{1}{c|}{3,907}         & \multicolumn{1}{c|}{2,342}         & \multicolumn{1}{c|}{739}          & 116          & 7,104                          \\ \hline
\multicolumn{5}{|c|}{\textit{Total}}                                                                                                                                                            & \textit{30,958}                         \\ \hline
\end{tabular}
\end{table}

\textbf{Method settings.} Our model for feature extraction uses several backbones for $f_e(;\theta)$, including ResNet18, VGG16, MobileNet-V2, and Inception-V3. These backbones are already pretrained on ImageNet22K. We use four data sources from each training dataset ($\mathsf{D}_{train}$) and the remaining data sources for $\mathsf{D}_{test}$. The learning rate $\eta$ is 0.001, the weight parameter $\gamma$ is 0.7, and there are four defect classes ($C$). The image will be resized to $64x64$ for classification purposes.
We train the classification models in a batch size of 32 for both $B_{te}$ and $B_{tr}$ in $N_{epoch}=20$ epochs using Adam optimizer and a learning rate of $1e^{-4}$. For our classification models, we set $\lambda$ to 1.0 and temperature parameter $\tau$ to 0.05. All experiments were performed on an NVIDIA Tesla V100-SXM3-32GB GPU device with 16GB of RAM, with the PyTorch library.

\textbf{Evaluation setting.} We use three standard metrics to benchmark classification models: precision (P), recall (R), and F1-score (F). We compare our method to several recent transfer learning methods for classifying road surface defects. The compared methods are Hsieh \etal \cite{method_raveling}, Hsieh \etal \cite{defect_convnet@Hsieh_2021}, Silva \etal \cite{defect_vgg16@Silva_2018}, Ali \etal \cite{defect_transfer_learning@Ali_2021}, Hasanaath \etal \cite{method_transfer_learning}, Chen \etal \cite{defect_mobilenet@Chen_2023}. These methods utilized a backbone pretrained on ImageNet for transfer learning. To validate the model's performance, we divided the data set into training and testing purposes. In each experiment, we use one data source as an unseen test set and the remaining five data sources for training. The performance of all compared models will be reported on unseen test set.

\subsection{Results}
The table \ref{tab:main_result} presents detailed results of our method and recent previous methods testing on six unseen data set of RDD2022. The results show that all methods perform less accurately on an unseen dataset. It demonstrates that the models are difficult to work well without training on an unseen dataset. It is reasonable because the unseen test set contains a road defect that differs significantly from the training set due to differences in road architecture across countries. The results also show that the model with MobileNet and ResNet backbones is more stable on an unseen dataset than the VGG backbone. Compared to other methods, our method has significantly higher accuracy in all unseen test sets. Furthermore, all of the performance metrics in all of the tested backbones of our method outperform previous methods by a large margin. This result shows that our method is more generalizable than the previous method for classifying road surface defects. 
\begin{table*}[ht]
\caption{Comparison results between our method and several recent methods with diverse backbones on unseen testing data from RDD2022.}
\label{tab:main_result}
\resizebox{\textwidth}{!}{%
\begin{tabular}{|l|l|cccccccccccccccccc|}
\hline
\multicolumn{1}{|c|}{\multirow{3}{*}{\textbf{Method}}} & \multicolumn{1}{c|}{\multirow{3}{*}{\textbf{Backbone}}} & \multicolumn{18}{c|}{\textbf{Unseen testing data source}}                                                                                                                                                                                                                                                                                                                                                                                                                                                                                                                                                                                                          \\ \cline{3-20} 
\multicolumn{1}{|c|}{}                                 & \multicolumn{1}{c|}{}                                   & \multicolumn{3}{c|}{\textbf{China MotorBike}}                                                                & \multicolumn{3}{c|}{\textbf{Czech}}                                                                          & \multicolumn{3}{c|}{\textbf{India}}                                                                          & \multicolumn{3}{c|}{\textbf{Japan}}                                                                          & \multicolumn{3}{c|}{\textbf{Norway}}                                                                         & \multicolumn{3}{c|}{\textbf{United States}}                                             \\ \cline{3-20} 
\multicolumn{1}{|c|}{}                                 & \multicolumn{1}{c|}{}                                   & \multicolumn{1}{c|}{\textbf{P}}    & \multicolumn{1}{c|}{\textbf{R}}    & \multicolumn{1}{c|}{\textbf{F}}    & \multicolumn{1}{c|}{\textbf{P}}    & \multicolumn{1}{c|}{\textbf{R}}    & \multicolumn{1}{c|}{\textbf{F}}    & \multicolumn{1}{c|}{\textbf{P}}    & \multicolumn{1}{c|}{\textbf{R}}    & \multicolumn{1}{c|}{\textbf{F}}    & \multicolumn{1}{c|}{\textbf{P}}    & \multicolumn{1}{c|}{\textbf{R}}    & \multicolumn{1}{c|}{\textbf{F}}    & \multicolumn{1}{c|}{\textbf{P}}    & \multicolumn{1}{c|}{\textbf{R}}    & \multicolumn{1}{c|}{\textbf{F}}    & \multicolumn{1}{c|}{\textbf{P}}    & \multicolumn{1}{c|}{\textbf{R}}    & \textbf{F}    \\ \hline
Hsieh \etal \cite{method_raveling}                          & ResNet18                           & \multicolumn{1}{c|}{0.43}          & \multicolumn{1}{c|}{0.46}          & \multicolumn{1}{c|}{0.43}          & \multicolumn{1}{c|}{0.41}          & \multicolumn{1}{c|}{0.44}          & \multicolumn{1}{c|}{0.38}          & \multicolumn{1}{c|}{0.39}          & \multicolumn{1}{c|}{0.42}          & \multicolumn{1}{c|}{0.27}          & \multicolumn{1}{c|}{0.41}          & \multicolumn{1}{c|}{0.4}           & \multicolumn{1}{c|}{0.38}          & \multicolumn{1}{c|}{0.36}          & \multicolumn{1}{c|}{0.4}           & \multicolumn{1}{c|}{0.35}          & \multicolumn{1}{c|}{0.37}          & \multicolumn{1}{c|}{0.43}          & 0.36          \\ \hline
Hsieh \etal \cite{defect_convnet@Hsieh_2021}                          & ResNet50                           & \multicolumn{1}{c|}{0.44}          & \multicolumn{1}{c|}{0.46}          & \multicolumn{1}{c|}{0.45}          & \multicolumn{1}{c|}{0.45}          & \multicolumn{1}{c|}{0.48}          & \multicolumn{1}{c|}{0.42}          & \multicolumn{1}{c|}{0.38}          & \multicolumn{1}{c|}{0.4}           & \multicolumn{1}{c|}{0.28}          & \multicolumn{1}{c|}{0.38}          & \multicolumn{1}{c|}{0.37}          & \multicolumn{1}{c|}{0.35}          & \multicolumn{1}{c|}{0.33}          & \multicolumn{1}{c|}{0.36}          & \multicolumn{1}{c|}{0.31}          & \multicolumn{1}{c|}{0.39}          & \multicolumn{1}{c|}{0.41}          & 0.35          \\ \hline
Silva \etal \cite{defect_vgg16@Silva_2018}                          & VGG16                              & \multicolumn{1}{c|}{0.4}           & \multicolumn{1}{c|}{0.44}          & \multicolumn{1}{c|}{0.42}          & \multicolumn{1}{c|}{0.38}          & \multicolumn{1}{c|}{0.4}           & \multicolumn{1}{c|}{0.35}          & \multicolumn{1}{c|}{0.39}          & \multicolumn{1}{c|}{0.41}          & \multicolumn{1}{c|}{0.28}          & \multicolumn{1}{c|}{0.38}          & \multicolumn{1}{c|}{0.36}          & \multicolumn{1}{c|}{0.35}          & \multicolumn{1}{c|}{0.32}          & \multicolumn{1}{c|}{0.36}          & \multicolumn{1}{c|}{0.31}          & \multicolumn{1}{c|}{0.36}          & \multicolumn{1}{c|}{0.4}           & 0.33          \\ \hline
Ali \etal \cite{defect_transfer_learning@Ali_2021}                          & VGG19                              & \multicolumn{1}{c|}{0.42}          & \multicolumn{1}{c|}{0.45}          & \multicolumn{1}{c|}{0.41}          & \multicolumn{1}{c|}{0.4}           & \multicolumn{1}{c|}{0.42}          & \multicolumn{1}{c|}{0.37}          & \multicolumn{1}{c|}{0.39}          & \multicolumn{1}{c|}{0.42}          & \multicolumn{1}{c|}{0.3}           & \multicolumn{1}{c|}{0.4}           & \multicolumn{1}{c|}{0.39}          & \multicolumn{1}{c|}{0.37}          & \multicolumn{1}{c|}{0.33}          & \multicolumn{1}{c|}{0.38}          & \multicolumn{1}{c|}{0.33}          & \multicolumn{1}{c|}{0.38}          & \multicolumn{1}{c|}{0.43}          & 0.34          \\ \hline
Hasanaath \etal \cite{method_transfer_learning}                         & MobileNet-V2                       & \multicolumn{1}{c|}{0.43}          & \multicolumn{1}{c|}{0.44}          & \multicolumn{1}{c|}{0.42}          & \multicolumn{1}{c|}{0.38}          & \multicolumn{1}{c|}{0.39}          & \multicolumn{1}{c|}{0.35}          & \multicolumn{1}{c|}{0.39}          & \multicolumn{1}{c|}{0.4}           & \multicolumn{1}{c|}{0.26}          & \multicolumn{1}{c|}{0.39}          & \multicolumn{1}{c|}{0.38}          & \multicolumn{1}{c|}{0.36}          & \multicolumn{1}{c|}{0.36}          & \multicolumn{1}{c|}{0.38}          & \multicolumn{1}{c|}{0.34}          & \multicolumn{1}{c|}{0.38}          & \multicolumn{1}{c|}{0.41}          & 0.34          \\ \hline
Chen \etal \cite{defect_mobilenet@Chen_2023}                         & MobileNet-V3                       & \multicolumn{1}{c|}{0.43}          & \multicolumn{1}{c|}{0.45}          & \multicolumn{1}{c|}{0.43}          & \multicolumn{1}{c|}{0.38}          & \multicolumn{1}{c|}{0.41}          & \multicolumn{1}{c|}{0.35}          & \multicolumn{1}{c|}{0.35}          & \multicolumn{1}{c|}{0.39}          & \multicolumn{1}{c|}{0.28}          & \multicolumn{1}{c|}{0.39}          & \multicolumn{1}{c|}{0.4}           & \multicolumn{1}{c|}{0.37}          & \multicolumn{1}{c|}{0.36}          & \multicolumn{1}{c|}{0.38}          & \multicolumn{1}{c|}{0.34}          & \multicolumn{1}{c|}{0.39}          & \multicolumn{1}{c|}{0.44}          & 0.36          \\ \hline
Ali \etal \cite{defect_transfer_learning@Ali_2021}                         & Inception-V3                       & \multicolumn{1}{c|}{0.41}          & \multicolumn{1}{c|}{0.43}          & \multicolumn{1}{c|}{0.41}          & \multicolumn{1}{c|}{0.4}           & \multicolumn{1}{c|}{0.42}          & \multicolumn{1}{c|}{0.38}          & \multicolumn{1}{c|}{0.36}          & \multicolumn{1}{c|}{0.41}          & \multicolumn{1}{c|}{0.3}           & \multicolumn{1}{c|}{0.4}           & \multicolumn{1}{c|}{0.4}           & \multicolumn{1}{c|}{0.37}          & \multicolumn{1}{c|}{0.35}          & \multicolumn{1}{c|}{0.38}          & \multicolumn{1}{c|}{0.35}          & \multicolumn{1}{c|}{0.39}          & \multicolumn{1}{c|}{0.44}          & 0.37          \\ \hline
\multirow{4}{*}{\textbf{Ours.}}    & ResNet-18                          & \multicolumn{1}{c|}{\textbf{0.5}}  & \multicolumn{1}{c|}{\textbf{0.53}} & \multicolumn{1}{c|}{\textbf{0.5}}  & \multicolumn{1}{c|}{\textbf{0.48}} & \multicolumn{1}{c|}{\textbf{0.5}}  & \multicolumn{1}{c|}{\textbf{0.46}} & \multicolumn{1}{c|}{\textbf{0.45}} & \multicolumn{1}{c|}{\textbf{0.5}}  & \multicolumn{1}{c|}{\textbf{0.35}} & \multicolumn{1}{c|}{\textbf{0.47}} & \multicolumn{1}{c|}{\textbf{0.47}} & \multicolumn{1}{c|}{\textbf{0.44}} & \multicolumn{1}{c|}{\textbf{0.45}} & \multicolumn{1}{c|}{\textbf{0.48}} & \multicolumn{1}{c|}{\textbf{0.44}} & \multicolumn{1}{c|}{\textbf{0.43}} & \multicolumn{1}{c|}{\textbf{0.5}}  & \textbf{0.43} \\ \cline{2-20} 
                                 & VGG16                              & \multicolumn{1}{c|}{\textbf{0.48}} & \multicolumn{1}{c|}{\textbf{0.5}}  & \multicolumn{1}{c|}{\textbf{0.48}} & \multicolumn{1}{c|}{\textbf{0.45}} & \multicolumn{1}{c|}{\textbf{0.44}} & \multicolumn{1}{c|}{\textbf{0.43}} & \multicolumn{1}{c|}{\textbf{0.44}} & \multicolumn{1}{c|}{\textbf{0.47}} & \multicolumn{1}{c|}{\textbf{0.33}} & \multicolumn{1}{c|}{\textbf{0.44}} & \multicolumn{1}{c|}{\textbf{0.43}} & \multicolumn{1}{c|}{\textbf{0.41}} & \multicolumn{1}{c|}{\textbf{0.4}}  & \multicolumn{1}{c|}{\textbf{0.43}} & \multicolumn{1}{c|}{\textbf{0.38}} & \multicolumn{1}{c|}{\textbf{0.41}} & \multicolumn{1}{c|}{\textbf{0.46}} & \textbf{0.4}  \\ \cline{2-20} 
                                 & MobileNet-V2                       & \multicolumn{1}{c|}{\textbf{0.51}} & \multicolumn{1}{c|}{\textbf{0.53}} & \multicolumn{1}{c|}{\textbf{0.49}} & \multicolumn{1}{c|}{\textbf{0.43}} & \multicolumn{1}{c|}{\textbf{0.44}} & \multicolumn{1}{c|}{\textbf{0.42}} & \multicolumn{1}{c|}{\textbf{0.44}} & \multicolumn{1}{c|}{\textbf{0.49}} & \multicolumn{1}{c|}{\textbf{0.34}} & \multicolumn{1}{c|}{\textbf{0.45}} & \multicolumn{1}{c|}{\textbf{0.44}} & \multicolumn{1}{c|}{\textbf{0.42}} & \multicolumn{1}{c|}{\textbf{0.44}} & \multicolumn{1}{c|}{\textbf{0.47}} & \multicolumn{1}{c|}{\textbf{0.42}} & \multicolumn{1}{c|}{\textbf{0.45}} & \multicolumn{1}{c|}{\textbf{0.49}} & \textbf{0.44} \\ \cline{2-20} 
                                 & Inception-V3                       & \multicolumn{1}{c|}{\textbf{0.49}} & \multicolumn{1}{c|}{\textbf{0.5}}  & \multicolumn{1}{c|}{\textbf{0.47}} & \multicolumn{1}{c|}{\textbf{0.46}} & \multicolumn{1}{c|}{\textbf{0.45}} & \multicolumn{1}{c|}{\textbf{0.43}} & \multicolumn{1}{c|}{\textbf{0.42}} & \multicolumn{1}{c|}{\textbf{0.47}} & \multicolumn{1}{c|}{\textbf{0.32}} & \multicolumn{1}{c|}{\textbf{0.46}} & \multicolumn{1}{c|}{\textbf{0.45}} & \multicolumn{1}{c|}{\textbf{0.43}} & \multicolumn{1}{c|}{\textbf{0.43}} & \multicolumn{1}{c|}{\textbf{0.47}} & \multicolumn{1}{c|}{\textbf{0.42}} & \multicolumn{1}{c|}{\textbf{0.44}} & \multicolumn{1}{c|}{\textbf{0.48}} & \textbf{0.43} \\ \hline
\end{tabular}
}
\end{table*}

\textbf{Ablation study.}
\begin{table}[ht]
\centering
\caption{The ablation study yields results with average F1-score of all unseen test sets on various training losses in our method.}
\label{tab:ablation}
\begin{tabular}{|l|c|c|c|c|}
\hline
\multicolumn{1}{|c|}{\textbf{Backbone}} & $\mathcal{L}_{ce}$ & $\mathcal{L}_{ce},\mathcal{L}_{ct}$ & $\mathcal{L}_{ce},\mathcal{L}_{dg}$ & $\mathcal{L}_{ce},\mathcal{L}_{ct},\mathcal{L}_{dg}$ \\ \hline
ResNet-18         & 0.36                             & 0.39                                 & 0.4                                   & 0.43                                       \\ \hline
VGG16             & 0.34                             & 0.36                                 & 0.38                                  & 0.4                                        \\ \hline
MobileNet-V2      & 0.35                             & 0.38                                 & 0.39                                  & 0.42                                       \\ \hline
Inception-V3      & 0.36                             & 0.37                                 & 0.38                                   & 0.41                                       \\ \hline
\end{tabular}
\end{table}
The table \ref{tab:ablation} shows the ablation study results for each metric learned in our method. We analyze the impact of training on multiple invariance data sources using contrastive learning separately from training domain generalization. Our method training with only traditional classification training loss. $\mathcal{L}_{ce}$ produces similar results to the previous method as shown in Table \ref{tab:main_result}. Our method improves model performance significantly by incorporating contrastive learning via $\mathcal{L}_{ct}$ (\textit{i.e.,} $\mathcal{L}_{ce},\mathcal{L}_{ct}$) or domain generalization loss via $\mathcal{L}_{dg}$ (\textit{i.e.,} $\mathcal{L}_{ce},\mathcal{L}_{dg}$) separately. When testing on an unseen test set, domain generalization training achieves slightly higher accuracy than training on multiple invariance data sources. Finally, our method, which incorporates both contrastive learning and domain generalization (\textit{i.e.,}$\mathcal{L}_{ce},\mathcal{L}_{ct},\mathcal{L}_{dg}$), achieves the highest average F1-score on all unseen test sets. This result shows that our method effectively addresses both the invariance data source and generalization challenges in road surface defect classification.

%%%%%%%%%%%%%%%%%%%%%%%%%%%%%%%%%%%%%%%%%%%%%%%%%%%%%%%%%%%%%%%%%%%%%%%%%%%%%%%%
\section{CONCLUSIONS}\label{sec:conclude}
In this study, we present a method for classifying road surface defects using camera images. There are two main parts to our method. Firstly, we propose an improvement to model training based on contrastive learning to address the invariance issue when developing a model using multiple datasets. Secondly, we propose an algorithm for training a generalized model that can be used with new unseen datasets without training the model on them. In the experiment, we simulate multiple data sources, including six data sets from six countries from the RDD2022 data set for road surface defects. The experiment results show that when compared to existing methods, our method is more efficient and accurate on the unsceen dataset. This indicates that our proposed method is well suited for generalizing models for road surface defect classification.
%%%%%%%%%%%%%%%%%%%%%%%%%%%%%%%%%%%%%%%%%%%%%%%%%%%%%%%%%%%%%%%%%%%%%%%%%%%%%%%%
\section*{ACKNOWLEDGMENT}
This work was realized in imec.ICON Hybrid AI for predictive road maintenance (HAIROAD) project, with the financial support of Flanders Innovation \& Entrepreneurship (VLAIO, project no. HBC.2023.0170).

%%%%%%%%%%%%%%%%%%%%%%%%%%%%%%%%%%%%%%%%%%%%%%%%%%%%%%%%%%%%%%%%%%%%%%%%%%%%%%%%

\bibliographystyle{IEEEtran}
\bibliography{reference}

\end{document}